\documentclass[sigconf,balance=false]{acmart}
\AtBeginDocument{%
  }

\usepackage{booktabs}
\usepackage{tabularx}
\usepackage{array}
\usepackage{soul}
\usepackage{graphicx}
\usepackage{amsmath}
\usepackage{multirow}
\usepackage{xcolor}

\copyrightyear{2026}
\acmYear{2026}
\setcopyright{cc}
\setcctype{by}
\acmConference[ICCAD '26]{IEEE/ACM International Conference on Computer-Aided Design}{November 08--12, 2026}{San Jose, CA, USA}
\acmBooktitle{IEEE/ACM International Conference on Computer-Aided Design (ICCAD '26), November 08--12, 2026, San Jose, CA, USA}
\acmDOI{10.1145/3831252.3847810}
\acmISBN{979-8-4007-2873-0/2026/11}

\begin{document}
\title{Can Agents Design Better Chips with a Higher Level Abstraction?}
\subtitle{(Invited Paper)}

\author{Zijian Ding}
\affiliation{%
  \institution{UCLA}
  \city{Los Angeles}
  \state{CA}
  \country{USA}
}
\email{bradyd@cs.ucla.edu}

\author{Yang Zou}
\affiliation{%
  \institution{UCLA}
  \city{Los Angeles}
  \state{CA}
  \country{USA}
}
\email{willz0123@g.ucla.edu}

\author{Yizhou Sun}
\affiliation{%
  \institution{UCLA}
  \city{Los Angeles}
  \state{CA}
  \country{USA}
}
\email{yzsun@cs.ucla.edu}

\author{Jason Cong}
\affiliation{%
  \institution{UCLA}
  \city{Los Angeles}
  \state{CA}
  \country{USA}
}
\email{cong@cs.ucla.edu}

\begin{abstract}
Large Language Model (LLM) agents are increasingly being explored for chip design, but most existing approaches operate directly at RTL. We ask whether agents can design better chips by leveraging higher-level abstractions. We compare Direct RTL Design, Agent-based HLS Design, Post-Compiler HLS Refinement, and Post-HLS RTL Refinement, and combine Agent-based HLS Design with Post-HLS RTL Refinement as Agent-based HLS with RTL Refinement (AHRR). We use FPGAs as a practical, easy-to-deploy platform for end-to-end evaluation, but note that the design-flow tradeoffs we study are largely independent of the target technology. Across a diverse 11-tasks benchmark suite, AHRR achieves a 2.6$\times$ geometric-mean speedup over Direct RTL Design across our benchmark suite. Case studies show that HLS distills design knowledge into abstractions that agents can leverage, while RTL refinement recovers lower-level optimization opportunities. Together, these results make AHRR a promising workflow for agentic chip design. The code and evaluation artifacts are available at https://github.com/ZijD/AHRR.
\end{abstract}



\maketitle

\section{Introduction}
\begin{figure*}[t]
    \centering
    \includegraphics[width=0.95\textwidth]{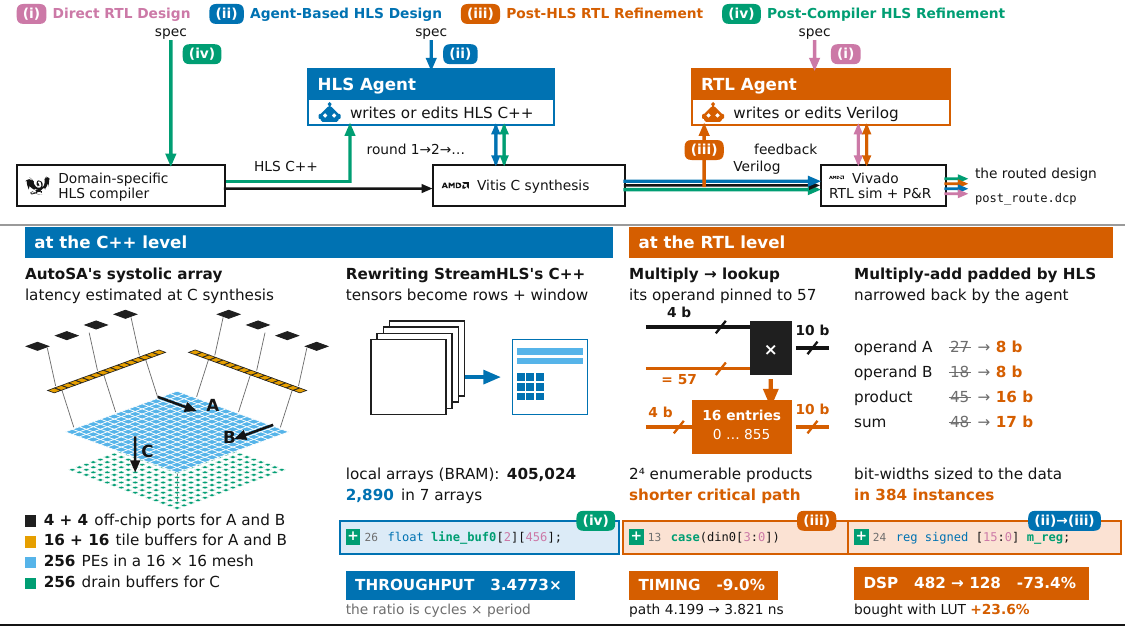}
    \caption{\small The four design flows studied in this paper. In (i) Direct RTL Design, the agent writes RTL directly from the specification. In (ii) Agent-based HLS Design, the agent writes HLS C++ and Vitis HLS generates RTL. In (iii) Post-HLS RTL Refinement, the agent further refines HLS-generated RTL. In (iv) Post-Compiler HLS Refinement, a domain-specific compiler generates the initial HLS C++, which the agent further refines.}
    \Description{Flow Diagram}
    \label{fig:flow}
\end{figure*}

Recent advances in Large Language Models (LLMs) and LLM-based coding agents have created growing interest in applying them to chip design. LLM agents have been explored for hardware generation, debugging, verification and optimization~\cite{verilogeval,cvdp,verilogcoder,mage,fvdebug}, and it is already applied in commercial design~\cite{jalapeño}.

However, RTL is not the only interface between an agent and the hardware design process. It has been widely used in manual designs as it provides cycle-accurate specification. In contrast, High-Level Synthesis (HLS)~\cite{cong2011high,cong2022fpga} lets designers express computation in C/C++ while delegating scheduling, pipelining, and RTL generation to a compiler, and domain-specific HLS frameworks~\cite{autosa,soda,streamhls} raise the abstraction further by providing and optimizing parameterized architecture templates. Agents have begun to work at the HLS level as well~\cite{agrefactor_arxiv,llmdse,lift}, but whether the abstraction itself helps an agent to produce a higher quality design is less clear. Direct RTL generation gives the agent full control, whereas HLS and domain-specific compilers restrict the design space to a usually high-quality subset. The two approaches can also be combined in multiple ways: (i) Direct RTL Design, where an agent creates RTL directly, (ii) Agent-based HLS Design, where an agent optimized HLS C/C++ code and then relies on HLS tools to generate RTL, (iii) Post-HLS RTL Refinement, where an agent further optimizes the HLS-generated RTL, and (iv) Post-Compiler HLS Refinement, where a domain-specific compiler first generates an HLS design that is further refined by an agent, with the resulting RTL serving as the input to flow (iii). We refer to flows (ii)+(iii) as AHRR: Agent-based HLS with RTL Refinement (Figure~\ref{fig:ahrr}), where the agent first designs at the HLS level and then refines the generated RTL.

These alternatives points to a fundamental question: \emph{can agents design better chips when operating at a higher level of abstraction using HLS, and how?}

Our contributions are:
\begin{itemize}
\item We systematically study LLM-based chip-design agents across RTL, HLS, and compiler-assisted design flows.
\item We show our proposed AHRR outperform Direct RTL by $2.6\times$ geometric-mean.
\item We further study Post-Compiler HLS Refinement, showing that agents can improve both HLS-generated RTL and compiler-generated HLS.
\end{itemize}

Evaluation across these design flows (Figure~\ref{fig:flow}) show that agents can design better chips by leveraging HLS and domain-specific HLS compilers that encode human design knowledge, combining machine and human intelligence.

\section{Background}
\label{sec:prelim}

\begin{table*}[t]
\centering
\small
\setlength{\tabcolsep}{3pt}
\caption{\small The 11-task benchmark suite spans LLM inference,
agentic memory, robotics, and GEMM. Selected tasks use
fixed-point arithmetic to avoid floating-point IP. Dynamic dimensions
are preserved when possible, and some workloads are scaled down to
reduce RTL simulation time.}
\label{tab:tasks}
\begin{tabular}{@{}>{\raggedright\arraybackslash}p{0.18\linewidth} >{\raggedright\arraybackslash}p{0.25\linewidth} >{\raggedright\arraybackslash}p{0.08\linewidth} >{\raggedright\arraybackslash}p{0.08\linewidth} >{\raggedright\arraybackslash}p{0.09\linewidth} >{\raggedright\arraybackslash}p{0.2\linewidth}@{}}
\toprule
Task & Category & Interface & Arithmetic & Dynamic axis & Scaled down \\
\midrule
\textbf{LLM1}\, \texttt{llm\_\allowbreak decode\_\allowbreak attention\_\allowbreak kv} & LLM decode attention over a KV cache~\cite{llama3} & \texttt{m\_\allowbreak axi} & fixed & \texttt{seq\_\allowbreak len} & no \\
\textbf{LLM2}\, \texttt{llm\_\allowbreak norm\_\allowbreak quant\_\allowbreak linear} & fused RMSNorm + quantize + INT8 linear~\cite{flexllm} & \texttt{m\_\allowbreak axi} & mixed & \texttt{seq\_\allowbreak len} & \textbf{yes} - 12 tokens \\
\textbf{GEM1}\, \texttt{gemm\_\allowbreak i8\_\allowbreak 512} & dense INT8 GEMM, two variants: 3/12 AXI & \texttt{m\_\allowbreak axi} & fixed & none & no \\
\textbf{MEM1}\, \texttt{mem\_\allowbreak bm25\_\allowbreak gather\_\allowbreak score} & BM25 retrieval ranking over postings~\cite{bm25} & \texttt{ap\_\allowbreak memory} & fixed & none & no \\
\textbf{MEM2}\, \texttt{mem\_\allowbreak kv\_\allowbreak page\_\allowbreak compact} & paged-KV~\cite{vllm} Prepare Memory stage & \texttt{ap\_\allowbreak memory} & fixed & \texttt{seq\_\allowbreak len} & no \\
\textbf{MEM3}\, \texttt{mem\_\allowbreak lserve\_\allowbreak page\_\allowbreak select} & LServe-style page selection~\cite{lserve} & \texttt{ap\_\allowbreak memory} & fixed & none & no \\
\textbf{ROB1}\, \texttt{robo\_\allowbreak nms\_\allowbreak boxes} & greedy non-maximum suppression & \texttt{ap\_\allowbreak memory} & fixed & \texttt{num\_\allowbreak boxes} & no \\
\textbf{ROB2}\, \texttt{robo\_\allowbreak pillar\_\allowbreak encoder} & PointPillars pillar feature encoder~\cite{pointpillars} & \texttt{ap\_\allowbreak memory} & fixed & \texttt{n\_\allowbreak points} & \textbf{yes} - coarser 0.25 m pillar \\
\textbf{ROB3}\, \texttt{robo\_\allowbreak stereo\_\allowbreak sad} & stereo block matching, SAD & \texttt{ap\_\allowbreak memory} & fixed & none & \textbf{yes} - IMG\_W/IMG\_H scaled down \\
\textbf{SHL1}\, \texttt{streamhls\_\allowbreak gemm} & float32 GEMM, generated by StreamHLS~\cite{streamhls} & \texttt{ap\_\allowbreak memory} & float & none & no \\
\textbf{SHL2}\, \texttt{streamhls\_\allowbreak dw\_\allowbreak conv} & dw-conv generated by StreamHLS & \texttt{ap\_\allowbreak memory} & float & none & no \\
\bottomrule
\end{tabular}
\end{table*}

HLS raises hardware design from RTL to C/C++ and reduces the verification burden~\cite{cong2011high,cong2022fpga}. Most agentic chip-design work targets
RTL~\cite{verilogeval,cvdp,verilogcoder,mage,thakur2024verigen}, while recent work explores HLS for pragma insertion, refactoring, and design-space exploration~\cite{hlspilot,collini2024c2hlsc,hlsrewriter,agrefactor_arxiv,llmdse,lift}. Domain-specific HLS compilers generate optimized architectures for particular computation patterns~\cite{autosa,soda,streamhls,ye2024hida,streamtensor,allo,scalehls,prometheus}. We use AutoSA and StreamHLS as starting points. Our agents follow the standard iterative coding-agent loop~\cite{react,reflexion,alphaevolve},
under a fixed round budget and a hidden testbench.

\section{Methodology}
\label{sec:method}

In this section, we describe the common agent loop, the four design
flows in Figure~\ref{fig:flow}, and the hidden testbench.

\paragraph{Specification-driven design.}
In Direct RTL Design (i) and Agent-based HLS Design (ii), the agent starts from the same task specification, task package, and evaluation harness. The task package includes a correct, unoptimized C++ implementation, one test stimulus, and a fixed interface contract. The agent then designs the accelerator either directly in RTL or in HLS C++. Within each round, the agent has up to one hour to develop and submit a candidate design. The evaluation harness records correctness on the hidden testbench, simulated cycle count, post-route clock period, and resource utilization. We use the open-source coding agent \texttt{pi}~\cite{pimono} without modification. The agent's session is carried across rounds, and each round's prompt reports the previous three rounds of evaluation results. We keep the prompt minimal, specifying only the optimization objective, resource budget, interface requirements, and submission protocol. In the RTL flow, the agent additionally receives the required port specification and, for AXI-based tasks, a provided AXI wrapper around the kernel.

\paragraph{Refinement from tool-generated designs.}
Flows (iii) and (iv) start from an existing implementation rather than only the task specification. In Post-Compiler HLS Refinement (iv), the agent starts from HLS C++ generated by a domain-specific compiler, such as AutoSA~\cite{autosa} or StreamHLS~\cite{streamhls}. In Post-HLS RTL Refinement (iii), it starts from Verilog produced by Vitis HLS, either from an Agent-based HLS design or a refined compiler-generated HLS design. Apart from the starting artifact and abstraction level, the optimization loop and evaluation procedure remain unchanged.

\emph{Hidden testbench.}
The public testbench provides a single stimulus for local development, while correctness is evaluated with a hidden testbench containing broader inputs and specification-derived corner cases. We strengthen each testbench until the C++ reference reaches $100\%$ branch coverage, and additionally measure statement and branch coverage on the reference RTL implementation.

\begin{table}[h]
\centering
\small
\setlength{\tabcolsep}{4pt}
\caption{\small Parameters fixed across experiments.}
\label{tab:params}
\begin{tabular}{@{}ll@{}}
\toprule
Parameter & Value \\
\midrule
device & AMD Alveo U55C (xcu55c-fsvh2892-2L-e) \\
toolchain & Vitis 2025.2, out of context (no shell) \\
agents & gemini-3.1-pro (high), gpt-5.6-sol (xhigh) \\
maximum rounds & 5 \\
agent round timeout & 60 min \\
evaluation timeout & RTL sim 75 min, syn + PnR 90 min \\
total timeout & 10 h \\
correctness check & C sim + RTL sim \\
clock target & 3.33 ns \\
achieved clock & max(target $-$ WNS, target) \\
resource ceiling & 60\% of BRAM, DSP, FF, LUT, URAM \\
\bottomrule
\end{tabular}
\end{table}

\section{Evaluation}
\label{sec:eval}

\begin{table*}[t]
\centering
\small
\setlength{\tabcolsep}{4pt}
\definecolor{ssgreen}{RGB}{0,158,115}

\caption{\small Comparison of the specification-driven design workflows.
Speedups are relative to Direct RTL Design on the same task and model.
Geometric means include only pairs with a valid Direct RTL baseline;
$n$ denotes the number of included pairs. AHRR has $n=13$ because
Post-HLS RTL Refinement was not evaluated on the two GEM1 (3 AXI)
cases.}
\label{tab:flowcompare}

\begin{tabular}{@{}ll
>{\raggedleft\arraybackslash}p{1.4cm}r
>{\raggedleft\arraybackslash}p{1.4cm}r
>{\raggedleft\arraybackslash}p{1.4cm}r@{}}
\toprule
& &
\multicolumn{2}{c}{(i) Direct RTL Design} &
\multicolumn{2}{c}{(ii) Agent-based HLS Design} &
\multicolumn{2}{c}{\begin{tabular}[b]{@{}c@{}}
(ii)$\rightarrow$(iii)\\AHRR
\end{tabular}} \\
\cmidrule(lr){3-4}
\cmidrule(lr){5-6}
\cmidrule(lr){7-8}

Task & Model &
ms & speedup &
ms & speedup &
ms & speedup \\
\midrule

\multirow{2}{*}{GEM1 (3 AXI)}
& gemini & 2.19 & 1.00$\times$
         & 0.40 & \textcolor{ssgreen}{5.43$\times$}
         & -- & -- \\
& gpt    & 1.31 & 1.00$\times$
         & 0.55 & \textcolor{ssgreen}{2.40$\times$}
         & -- & -- \\
\cmidrule{1-8}

\multirow{2}{*}{LLM1}
& gemini & 1.29 & 1.00$\times$
         & 0.43 & \textcolor{ssgreen}{3.02$\times$}
         & 0.43 & \textcolor{ssgreen}{3.02$\times$} \\
& gpt    & 2.54 & 1.00$\times$
         & 0.47 & \textcolor{ssgreen}{5.37$\times$}
         & 0.43 & \textcolor{ssgreen}{5.87$\times$} \\
\cmidrule{1-8}

\multirow{2}{*}{LLM2}
& gemini & $\times$ & 
         & 0.20 & --
         & 0.20 & -- \\
& gpt    & 1.23 & 1.00$\times$
         & 0.21 & \textcolor{ssgreen}{6.00$\times$}
         & 0.21 & \textcolor{ssgreen}{6.00$\times$} \\
\cmidrule{1-8}

\multirow{2}{*}{MEM1}
& gemini & $\times$ &
         & 7.44 & --
         & 7.44 & -- \\
& gpt    & 5.26 & 1.00$\times$
         & 5.27 & 1.00$\times$
         & 5.25 & \textcolor{ssgreen}{1.00$\times$} \\
\cmidrule{1-8}

\multirow{2}{*}{MEM2}
& gemini & $\times$ &
         & 0.37 & --
         & 0.37 & -- \\
& gpt    & 0.65 & 1.00$\times$
         & 0.37 & \textcolor{ssgreen}{1.75$\times$}
         & 0.32 & \textcolor{ssgreen}{2.02$\times$} \\
\cmidrule{1-8}

\multirow{2}{*}{MEM3}
& gemini & 3.62 & 1.00$\times$
         & 3.50 & \textcolor{ssgreen}{1.03$\times$}
         & 3.50 & \textcolor{ssgreen}{1.03$\times$} \\
& gpt    & 3.50 & 1.00$\times$
         & 3.50 & 1.00$\times$
         & 3.50 & 1.00$\times$ \\
\cmidrule{1-8}

\multirow{2}{*}{ROB1}
& gemini & 6.16 & 1.00$\times$
         & 0.50 & \textcolor{ssgreen}{12.35$\times$}
         & 0.45 & \textcolor{ssgreen}{13.82$\times$} \\
& gpt    & 0.30 & 1.00$\times$
         & 0.53 & 0.56$\times$
         & 0.33 & 0.90$\times$ \\
\cmidrule{1-8}

\multirow{2}{*}{ROB2}
& gemini & 35.38 & 1.00$\times$
         & 6.37 & \textcolor{ssgreen}{5.56$\times$}
         & 6.22 & \textcolor{ssgreen}{5.69$\times$} \\
& gpt    & 9.22 & 1.00$\times$
         & 2.61 & \textcolor{ssgreen}{3.53$\times$}
         & 2.20 & \textcolor{ssgreen}{4.19$\times$} \\
\cmidrule{1-8}

\multirow{2}{*}{ROB3}
& gemini & 0.02 & 1.00$\times$
         & 0.01 & \textcolor{ssgreen}{1.73$\times$}
         & 0.01 & \textcolor{ssgreen}{1.95$\times$} \\
& gpt    & 0.01 & 1.00$\times$
         & 0.01 & 0.54$\times$
         & 0.00 & \textcolor{ssgreen}{2.09$\times$} \\

\cmidrule(lr){2-8}
& \itshape geo.\ mean
& & 1.00$\times$ {\scriptsize($n$=15)}
& & 2.31$\times$ {\scriptsize($n$=15)}
& & 2.62$\times$ {\scriptsize($n$=13)} \\

\bottomrule
\end{tabular}
\end{table*}

In this section, we evaluate how the abstraction level affects agentic chip design. We first compare Direct RTL Design and Agent-based HLS Design to isolate the benefit of HLS. We then evaluate Post-HLS RTL Refinement and Post-Compiler HLS Refinement.

\subsection{Experimental Setup}
\label{sec:eval:setup}

We run Gemini 3.1 Pro at high reasoning effort and GPT-5.6-sol at extra-high effort under the same \texttt{pi} harness with every parameter fixed (Table~\ref{tab:params}).

A design is valid if it passes the hidden testbench, routes successfully, and stays within the resource ceiling. We compute execution time as the simulated cycle count multiplied by the maximum of the post-route clock period and the target clock period. All reported speedups are measured relative to execution time on the same task. Each experiment runs once. Table~\ref{tab:tasks} lists the eleven tasks. For GEM1, we have two variants, one with naive sequential design with 3 AXI masters (A,B,C) for RTL v.s. HLS comparison, and the other with generated by AutoSA with 12 AXI masters.

\subsection{Overview: Design-Flow Comparison}

Table~\ref{tab:flowcompare} compares the specification-driven
workflows. Agent-based HLS Design achieves a 2.31$\times$
geometric-mean speedup over Direct RTL Design. Extending it with
Post-HLS RTL Refinement to form AHRR increases the speedup to
2.62$\times$. We separately evaluate Post-Compiler HLS Refinement
in Section~\ref{sec:eval:part2}.

\subsection{Agent-based HLS vs. Direct RTL}
\label{sec:eval:part1}

We first isolate the effect of abstraction by giving the same agent the same task specification and asking it to design in either RTL or HLS, corresponding to flows (i) and (ii).

\begin{figure}[t]
    \centering
    \includegraphics[width=\columnwidth]{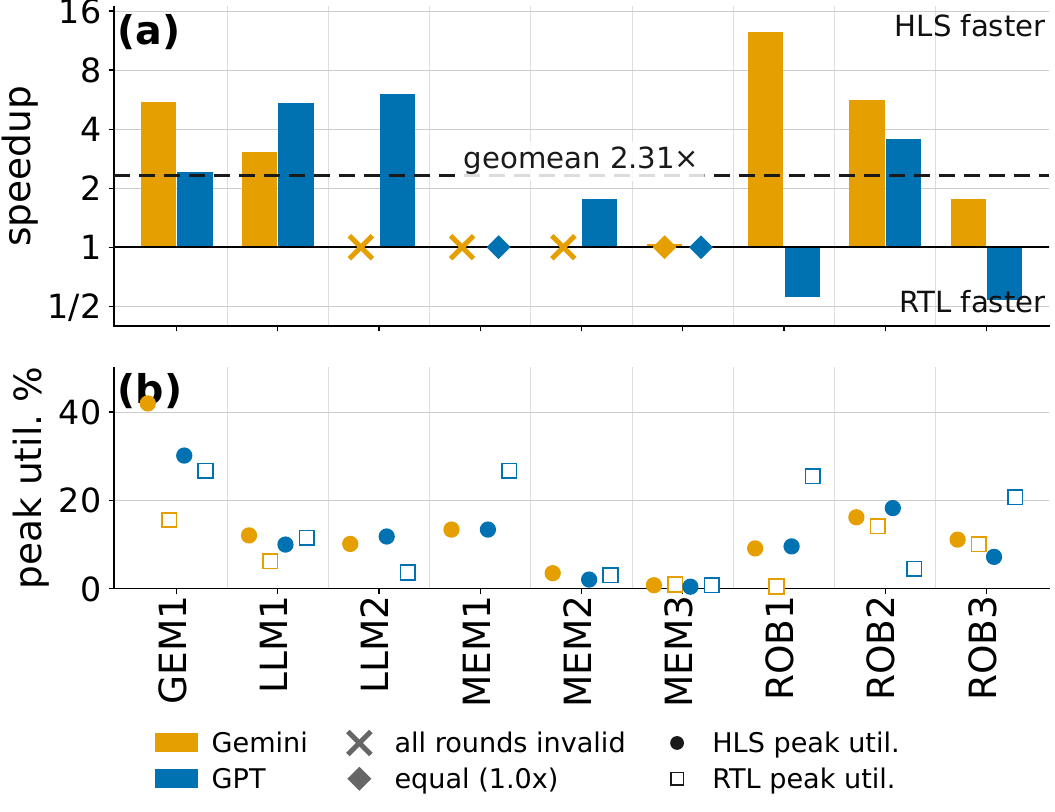}
    \caption{\small Performance and resource comparison of Agent-based HLS Design and Direct RTL Design. With HLS, state-of-the-art agents achieve a 2.31$\times$ geometric-mean speedup over direct RTL design.}
    \label{fig:hls_vs_rtl}
    \Description{Performance and resource comparison of Agent-based HLS Design and Direct RTL Design.}
\end{figure}

With the HLS tool available, the agent can express parallelism through loop transformations and pragmas. In RTL, the same agent must realize these structures explicitly and often produces a narrower datapath. This demonstrates that HLS tools carry distilled design knowledge that agents can leverage. We provide a detailed case study in Figure~\ref{fig:case_study_1}.

Direct RTL Design may still win when the design is small. The two cases where RTL outperforms HLS are both on small kernels, ROB1 and ROB3, and by 1.78$\times$ and 1.85$\times$ respectively. When the kernel state fits in registers, the RTL flow can place it there directly, while HLS can only express the intent through binding pragmas. We also note that we use naive agent for both Agent-based HLS Design and Direct RTL Design, and a simple prompt update may significantly improve/degrade performance on both sides.

We next ask whether the agent can combine both advantages by refining the HLS-generated RTL.

\subsection{Post-HLS RTL Refinement}
\label{sec:eval:postrtl}

We next apply Post-HLS RTL Refinement to the Verilog generated from the best Agent-based HLS design. The agent continues optimization directly at the RTL level.

Across the sixteen valid cases, Post-HLS RTL Refinement improves performance by 1.17$\times$ geometric mean over its HLS-generated RTL starting point.

\begin{figure*}[!h]
    \centering
    \includegraphics[width=0.85\textwidth]{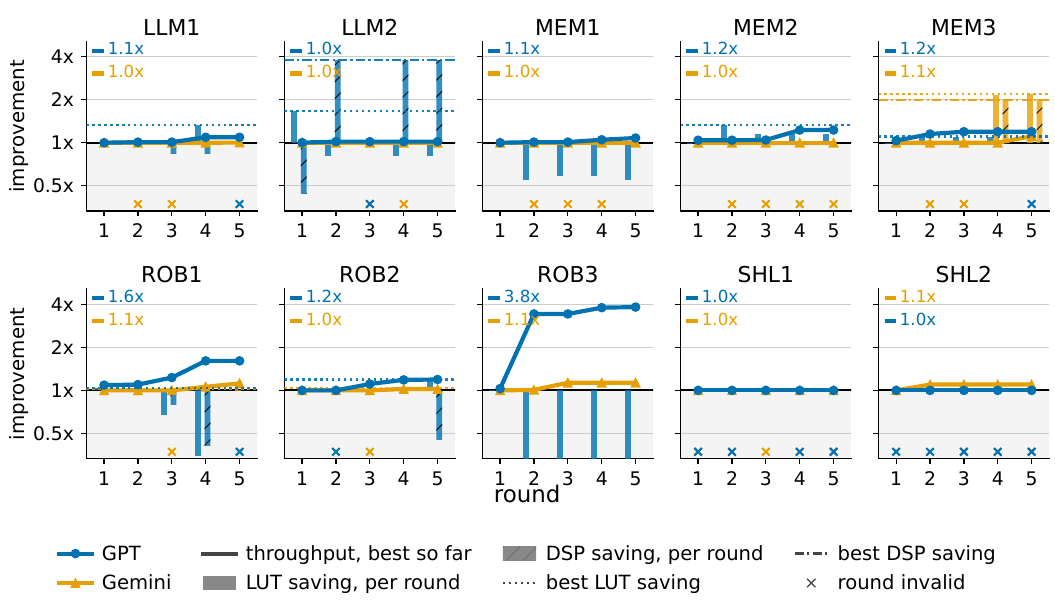}
    \caption{\small Per-round progress of Post-HLS RTL Refinement, normalized to the HLS-generated RTL that seeds each experiment. All axes are oriented so that higher is better. LUT and DSP savings are shown alongside speedup to capture the tradeoff between area and performance. We also include SHL1/SHL2 to illustrate that Post-HLS RTL Refinement becomes challenging when simulation time is long.}
    \label{fig:post_rtl}
    \Description{post hls rtl optimization}
\end{figure*}

Figure~\ref{fig:post_rtl} isolates the effect of Post-HLS RTL Refinement, showing the per-round progress of each agent. We use throughput/latency as the primary objective and resource usage as
the secondary objective. On LLM1, MEM1/2/3, and ROB2, agents make incremental improvements over the HLS-generated RTL. On ROB1 and ROB3, the agents achieve $2$-$4\times$ speedups through complete
rewrites, consistent with the results in Figure~\ref{fig:hls_vs_rtl}. On LLM2 and MEM3, GPT and Gemini also substantially reduce resource
usage with only a few lines of RTL edits, which we examine in Section~\ref{sec:cs:c}.

Most successful rounds make local edits rather than large rewrites, and these edits usually change the cycle count by less than 1.15$\times$. Every larger latency improvement comes from a complete rewrite, where the agent discards the HLS-generated RTL and writes a new implementation.

\subsection{Post-Compiler HLS Refinement}
\label{sec:eval:part2}

We finally study a stronger starting point, where a domain-specific compiler first generates the HLS design and the agent refines the generated C++. Table~\ref{tab:part2_postcompiler} reports the results. On SHL2, Gemini and GPT  improve StreamHLS by 3.3$\times$ and 1.7$\times$. On GEM1, both improve AutoSA by 1.3$\times$ and 1.2$\times$ while changing the cycle count by less than 2\%, with the gains coming from improved timing. Gemini changes storage bindings (''bind\_storage type=ram\_2p impl=lutram``) and FIFO depths (2->4), while GPT restructures control and the PE datapath. In both cases, the agents incrementally improve an already optimized compiler-generated HLS design, showing that useful optimization opportunities remain after compiler optimization.

\begin{table}[t]
\centering
\small
\setlength{\tabcolsep}{5pt}
\caption{\small Post-Compiler HLS Refinement relative to the
domain-specific compiler baseline, reported as Gemini / GPT.
We use StreamHLS for SHL1/SHL2 and AutoSA for GEM1.}
\label{tab:part2_postcompiler}

\begin{tabular}{lccc}
\toprule
& SHL1 & SHL2 & GEM1 \\
\midrule
DS-compiler
& StreamHLS & StreamHLS & AutoSA \\

Compiler baseline
& 1.0 & 1.0 & 1.0 \\

Post-Compiler HLS Refinement
& 1.0 / 1.0
& 3.3 / 1.7
& 1.3 / 1.2 \\
\bottomrule
\end{tabular}
\end{table}

For SHL1, neither agent improves upon the compiler-generated design. This is not necessarily an intrinsic limitation of the agents. The initial compiler-generated design already requires simulation time close to our timeout limit, leaving little room for iterative exploration. A better evaluation harness could use HLS latency estimates during optimization and reserve full simulation for validation, which we leave to future work.

\section{Case Study}
In this section, we examine the design and agent decisions that most clearly reveal the role of high-level abstractions in agentic chip design. We first show how Agent-based HLS Design can scale designs efficiently, illustrating how abstractions expose useful design knowledge. We then show that compilers translating these abstractions can still leave inefficiencies that agents can identify and improve. Finally, we examine how Post-HLS RTL Refinement can improve timing and area beneath the HLS-generated design.

\subsection{High Level Abstractions Distill Design Knowledge for Agent}
\label{sec:cs:a}

\begin{figure}[!h]
    \centering
    \includegraphics[width=\columnwidth]{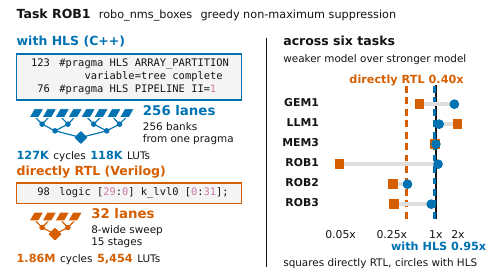}
    \caption{\small Abstractions distill design knowledge for agents. On the left, the same agent implements ROB1 through HLS and direct RTL Design. The HLS design uses two pragmas to produce 256-way parallelism, while the direct RTL design achieves only 32/8-way parallelism despite requiring more Verilog than HLS C++. On the right, across six tasks, the weaker model nearly matches the stronger model when both operate through HLS.}
    \label{fig:case_study_1}
    \Description{case study 1}
\end{figure}

Figure~\ref{fig:case_study_1} shows ROB1, greedy non-maximum suppression. Object detectors often produce multiple high-scoring, overlapping candidate boxes for the same object. Non-maximum suppression (NMS) reduces these duplicates by repeatedly selecting the highest-scoring remaining box and suppressing candidates whose intersection-over-union (IoU) with it exceeds a threshold~\cite{bodla2017soft}. ROB1 applies greedy NMS to up to 8192 candidate boxes, keeps at most 512, and uses a 50\% IoU threshold. For each retained box, the design performs an argmax pass to select the next box and a suppression pass to remove overlapping candidates. This requires roughly $2KN$ box visits for $K$ retained boxes among $N$ candidates, making the number of boxes processed per cycle the main latency bottleneck.

With HLS, the agent asked for that degree of parallelism in one pragma each, an \texttt{ARRAY\_PARTITION} into 256 banks and a \texttt{PIPELINE} at one iteration per cycle, and Vitis HLS supplied what the request implies: a 256-port read, a log-depth comparison tree and the pipeline registers between its levels, in 43914 lines of Verilog. The same agent writing Verilog directly built a 32-lane argmax and an 8-lane suppression sweep in 378 lines and 15 pipeline stages, reading one 512-bit word per cycle from the memory port. Its design runs 1.86M cycles against 127K, with 5454 LUTs against HLS-generated RTL's 118K. The right panel of Figure~\ref{fig:case_study_1} shows the same effect across models: on the six tasks where all 4 pairs of (GPT/Gemini)x(HLS/RTL) succeed, the weaker model reaches 0.40$\times$ the speedup of the stronger with direct RTL and 0.95$\times$ with HLS. This demonstrate that HLS simplify the effort of design optimization.

\subsection{Agents Recover Inefficiencies Left by Compiler Heuristics}
\label{sec:cs:b}

\begin{figure}[!h]
    \centering
    \includegraphics[width=\columnwidth]{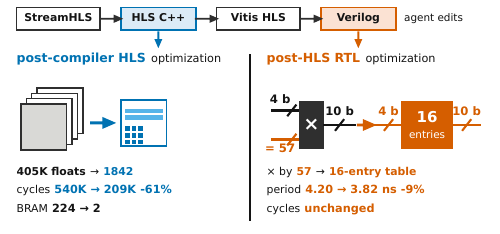}
    \caption{\small Agents recover inefficiencies left by compiler heuristics. The StreamHLS design passes through Post-Compiler HLS Refinement and Post-HLS RTL Refinement. At HLS level, Four whole tensors are rewritten to two rows and a window, reducing cycle count by 61\%. At RTL level, a multiplier with a pinned operand becomes a lookup table, reducing the clock period by 9\% at unchanged cycle count.}
    \label{fig:case_study_2}
    \Description{case study 2}
\end{figure}

Figure~\ref{fig:case_study_2} follows the SHL2 design through Post-Compiler HLS Refinement (iv) and Post-HLS RTL Refinement (iii). StreamHLS generates four local arrays totaling 405K floating-point values and adds an extra stage to fill a padded copy before the convolution reads it. This illustrates a common limitation of source-to-source compilers: even when the abstraction
points optimization in the right direction, code generation still relies on fixed heuristics that may not generalize across designs.

This is where agents can complement compiler automation. With Post-Compiler HLS Refinement, the agent kept two rows of line buffer and a 3$\times$3 window, 1842 floating points in all, and deleted the padding stage. The cycle count fell from 539726 to 209297 and the block RAM from 224 to 2.

The same design also benefits from Post-HLS RTL Refinement. The agent identifies two multiplier modules whose second operands are already constant in the generated RTL and replaces them with lookup tables. The clock period drops from 4.20 to 3.82 ns with no change in cycle count.

This case illustrates an ideal division of optimization, where compilers and higher-level abstractions make the ``giant steps'' and agents recover case-specific ``baby steps'' with surgical precision.

\subsection{Agents optimize timing and area beneath the HLS schedule}
\label{sec:cs:c}

\begin{figure}[!h]
    \centering
    \includegraphics[width=\columnwidth]{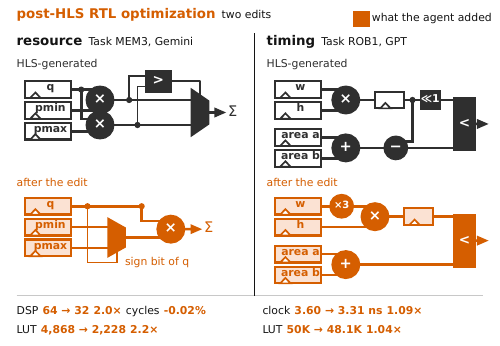}
    \caption{\small Agents optimize timing and area beneath the HLS schedule: on MEM3, moving the min/max selection before multiplication cuts DSP usage and cuts LUTs by 2.2$\times$ without changing the cycle count. On ROB1, rewriting the comparison removes a subtractor from the critical path and reduces the clock period from 3.60 to 3.31 ns.}
    \label{fig:case_study_3}
    \Description{case study 3}
\end{figure}

Figure~\ref{fig:case_study_3} traces two designs through Post-HLS RTL Refinement (iii). At the HLS level, both computations are expressed compactly, but the generated RTL exposes lower-level implementation choices that the agent can further improve.

On MEM3, the HLS design computes $\max(q\cdot\mathit{min}, q\cdot\mathit{max})$ for each of 32 vector elements. Vitis HLS implements this with two multipliers per element and selects the larger product, using 64 DSPs in total. Gemini observes that the sign of $q$ alone determines whether $\mathit{min}$ or $\mathit{max}$ should be used, and moves this selection before the multiplication. This reduces the design to one multiplier per element, cutting DSP usage from 64 to 32 and LUT usage from 4,868 to 2,228, while keeping the cycle count nearly unchanged. The clock period also improves from 3.14 to 2.83 ns.

On ROB1, the generated RTL implements the 50\% intersection-over-union comparison as
$\mathit{union} < 2\cdot\mathit{intersection}$, where
$\mathit{union}=a+b-\mathit{intersection}$. This places a multiplier, a 33-bit subtractor, and a comparator on the critical path. GPT instead rewrites the equivalent condition as
$a+b<3\cdot\mathit{intersection}$, removing the subtractor from the path. The clock period improves from 3.60 to 3.31 ns, meeting the 3.33 ns target, with the same cycle count and 4\% fewer LUTs.

These examples show another useful division of optimization. HLS captures the high-level computation and generates a complete implementation, while the agent can inspect the resulting RTL and recover case-specific opportunities in resource mapping, arithmetic structure, and timing that only become explicit after RTL generation.

\subsection{Recommended Flow}
\begin{figure}
    \centering
    \includegraphics[width=\columnwidth]{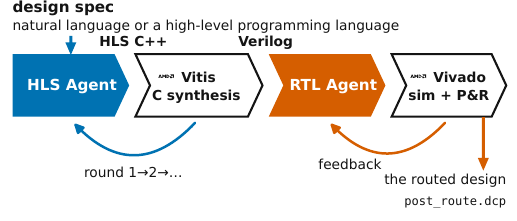}
    \caption{\small The proposed AHRR workflow.}
    \label{fig:ahrr}
\end{figure}

Our results suggest combining Agent-based HLS Design with Post-HLS RTL Refinement as a promising flow for agentic chip design, which means that starting starting with a design specification in natural language or some high-level executable language, an agent generates optimized HLS C/C++ code, and the further inspects the generated RTL code by an HLS tool for further refinement at the RTL level. We name it the AHRR Flow. Figure~\ref{fig:ahrr} illustrates the AHRR flow.

The key strength of AHRR is its combination of \emph{architectural leverage} from HLS and \emph{cycle-accurate control} at RTL. At the HLS level, the agent can efficiently explore broad architectural transformations through compact C/C++ changes and compiler directives. After RTL generation, the same agent can inspect the generated implementation and further refine timing, storage, and datapath decisions that only become explicit at the lower level. These two stages are complementary: Agent-based HLS Design already achieves a 2.31$\times$ geometric-mean speedup over Direct RTL Design, while subsequent RTL refinement increases the speedup to 2.62$\times$. This ability to operate effectively on both sides of the HLS compiler boundary makes AHRR particularly promising for agentic chip design.

\section{Conclusion}
\label{sec:conclusion}

In this study, through extensive evaluation using the latest commercial AI agents in conjunction with state-of-the-art HLS tools, we conclude that the AHRR flow proposed in the preceding section is particularly promising for high-quality RTL design generation. The key strength of this approach is its combination of \textbf{machine intelligence}, embodied in frontier AI models, and \textbf{human intelligence}, embodied in HLS and domain-specific compilers. HLS provides a correct-by-construction mechanism for bridging the gap between behavioral specifications expressed in C/C++ or other high-level languages and the cycle-accurate implementations required for RTL designs. Meanwhile, today’s coding agents have demonstrated remarkable capabilities in high-level programming, leveraging the vast amount of software code represented in their training data, as well as an impressive ability to understand and refine RTL generated by HLS compilers. The combination of these complementary capabilities makes the AHRR flow particularly powerful in the agentic era.

It is worth noting that, despite the significant productivity gains offered by HLS, HLS-based design methodologies have historically seen limited adoption, due in part to two barriers: (i) the inertia of existing RTL-based design practices, which makes designers reluctant to adopt a new methodology, and (ii) the difficulty for human designers to read and understand HLS-generated RTL when cycle-accurate modifications are required. AI agents can overcome both barriers. Unlike human designers, agents can readily explore and adopt HLS-based methodologies when they produce better results, without being constrained by the inertia of established design practices. Moreover, agents can effectively understand and further refine the RTL generated by HLS tools when low-level control is necessary. We therefore expect agents to substantially accelerate the adoption of HLS and other high-level design methodologies. More broadly, our results suggest a promising paradigm for chip design: \textbf{agents can design better chips by leveraging tools that operate at higher levels of abstraction.} A similar view was also expressed in a recent position paper~\cite{zhang2026pragmas}.

As we completed this study, it is encouraging to see that AI agent + HLS design is being adopted by some major industry players, such as the Jalapeño chip designed by OpenAI and announced in August 2026~\cite{jalapeño}. No details on the design flow has been released yet, but it is an exciting development.

\begin{acks}
This work was partially supported by NSF grants DGE-2034835, SRC JUMP 2.0 PRISM Center, Google, Jane Street, and the CDSC industrial partners (https://cdsc.ucla.edu/partners/). The authors thank AMD/Xilinx for HACC equipment donation and HPCfund cluster access. J.~Cong has a financial interest in AMD and Google.
\end{acks}

\bibliographystyle{ACM-Reference-Format}
\bibliography{main}

\end{document}